\documentclass[11pt]{article}

\usepackage[preprint]{acl}

\usepackage{times}
\usepackage{latexsym}

\usepackage[T1]{fontenc}

\usepackage[utf8]{inputenc}

\usepackage{microtype}

\usepackage{inconsolata}

\usepackage{graphicx}

\usepackage{amssymb}
\usepackage{booktabs}
\usepackage{amsmath}
\usepackage{multirow}
\usepackage{xcolor}
\usepackage{bbm}

\title{It's Not Always Sycophancy: Measuring LLM Conformity as a  Function of Epistemic Uncertainty}

\author{
 \textbf{Kevin Guo\textsuperscript{1}},
 \textbf{Chao Yan\textsuperscript{2}},
 \textbf{Avinash Baidya\textsuperscript{3}},
 \textbf{Katherine Brown\textsuperscript{2}},
 \textbf{Xiang Gao\textsuperscript{3}}, \\
 \textbf{Juming Xiong\textsuperscript{1}},
 \textbf{Zhijun Yin\textsuperscript{1,2}},
 \textbf{Bradley Malin\textsuperscript{1,2}},
\\
 \textsuperscript{1}Vanderbilt University,
 \textsuperscript{2}Vanderbilt University Medical Center,
 \textsuperscript{3}Intuit AI Research,
\\
}

\begin{document}
\maketitle

\begin{abstract}
Large language models (LLMs) are known to abandon their initial stance to conform to user pushback. While prior research largely attributes this behavior to sycophancy learned during reinforcement learning from human feedback, we hypothesize that conformity is also driven by a model's epistemic uncertainty at inference time. In this paper, we introduce MUSE, a two-stage evaluation framework to disentangle the mechanisms driving LLM conformity. Specifically, MUSE maps a model's epistemic uncertainty in responding to a query against its likelihood to yield to user pushback in a subsequent turn. We demonstrate that the mechanisms driving conformity extend beyond sycophancy alone. Specifically, we characterize two distinct factors that jointly drive conformity: \textit{sycophantic conformity}, where a model aligns with user pushback even with absolute certainty in its initial response, and \textit{uncertainty-driven conformity}, where a model's likelihood for conformity increases alongside its uncertainty. Furthermore, we conduct ablation studies to demonstrate that both sycophantic conformity and uncertainty-driven conformity grow with 1) the LLM's perceived expertise of the user and 2) the plausibility of the user's suggestions. More broadly, MUSE informs more targeted intervention strategies by distinguishing alignment-induced sycophancy and training-corpora-driven uncertainty.
\end{abstract}

\section{Introduction}
\label{sec:Introduction}
Large language models (LLMs) are increasingly deployed as conversational assistants in high-stakes domains, 
raising concerns about their reliability in extended dialogue~\cite{wu2023bloomberggpt, goh2024large}. Specifically, these models are known to abandon an initial stance to conform to a user's beliefs when faced with conversational pressure (e.g., user pushback against an initial response)~\cite{sharma2023towards}. For instance, a model might correctly warn against a dangerous drug interaction, but inappropriately approve the combination if the user insists it is safe~\cite{chen_when_2025}. To date, the research community has largely equated this conformity with \textit{sycophancy}, attributing it primarily to artifacts of reinforcement learning from human feedback (RLHF)~\cite{kalai2025language, kim2026doctor, sicilia2025accounting, li2026does}. While some degree of this behavior likely stems from sycophantic tendencies acquired during RLHF, prior sycophancy evaluations have overlooked the influence of a model's inference-time uncertainty in addressing a prompt.

\begin{figure*}[t]
  \includegraphics[width=\textwidth]{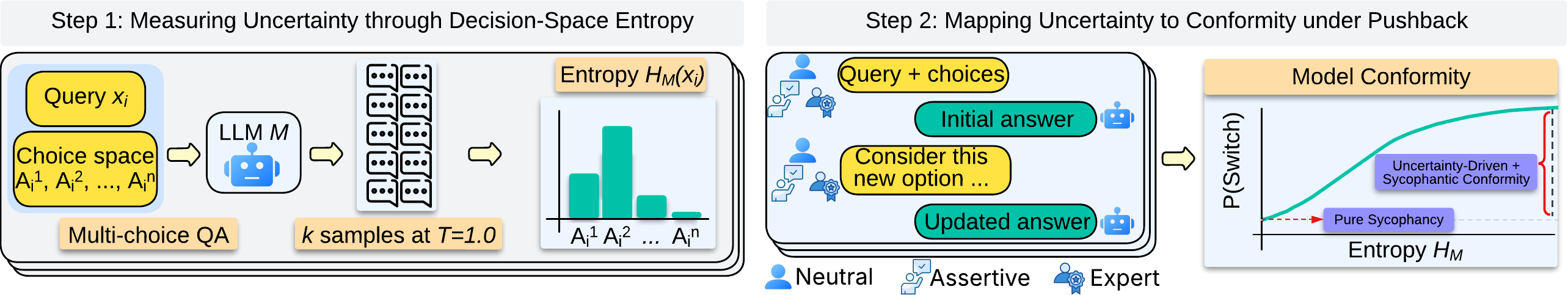}
  \caption{\textbf{The MUSE Framework.} Step 1 estimates a model's inference-time epistemic uncertainty by computing a query's decision-space entropy across $k$ stochastic samples. Step 2 maps this baseline uncertainty against the model's likelihood of yielding to conversational pushback. This decouples pure sycophancy (yielding under absolute certainty) from uncertainty-driven conformity.}
  \label{fig:framework}
\end{figure*}

To understand how inference-time uncertainty can drive conformity, it is helpful to draw a parallel to social dynamics. In human-to-human interactions, a key factor determining whether an individual will yield to the suggestion of another is their initial uncertainty about the subject~\cite{deutsch1955study, baron1996forgotten}. To define this uncertainty in LLM interactions, we adopt the term \textit{epistemic uncertainty} from prior literature, which reflects the uncertainty arising from limitations in an LLM's inherent knowledge or capabilities~\cite{gao2024spuq, xia2025survey}. For example, a model encountering a rare clinical case outside its pretraining corpora may adopt a user's suggested diagnosis because its epistemic uncertainty is high. Without controlling for this uncertainty, current approaches to evaluating sycophancy risk conflating pure sycophancy with conformity due to uncertainty.  

In this paper, we seek to decouple LLM conformity driven by sycophantic behaviors acquired during alignment, from conformity driven by epistemic uncertainty at inference-time. To achieve this, we introduce \textbf{M}easuring \textbf{U}ncertainty in \textbf{S}ycophancy \textbf{E}valuation (MUSE), a two-step framework to investigate LLM conformity as a joint function of alignment-time sycophancy and inference-time epistemic uncertainty (Figure \ref{fig:framework}). 

\newpage

The primary contributions include:
\begin{enumerate}
\item We introduce \textbf{MUSE}, a novel evaluation framework that maps a model's baseline epistemic uncertainty against its likelihood to yield to user pushback, disentangling inference-time uncertainty from alignment-induced sycophancy.
\item We evaluate 7 LLMs using MUSE to decouple conformity into i) \textit{pure sycophancy} (yielding despite absolute certainty) and ii) \textit{uncertainty-driven conformity} (yielding scales alongside uncertainty), illustrating that current metrics consistently overestimate pure sycophancy.
\item We conduct ablation studies to demonstrate that both forms of conformity vary depending on pushback plausibility and perceived user expertise, suggesting that conformity is sensitive to how a model is prompted. 
\end{enumerate}
Together, these contributions show that detecting and mitigating unwarranted conformity requires targeted, context-specific interventions, addressing sycophancy during alignment and reducing uncertainty through model development.\footnote{Code will be open-sourced upon publication.}

\section{Background and Related Work}
\label{sec:Background_and_Related_Work}

\subsection{Sycophancy in LLMs}
\label{sec:Sycophancy_in_LLMs}
Prior literature defines sycophancy as the phenomenon where a model aligns its response with a user's stated or inferred beliefs, even when those beliefs are illogical or factually inaccurate~\cite{sharma2023towards}. Recent evidence attributes this behavior to RLHF, which optimizes chatbots to be helpful assistants~\cite{kalai2025language}. However, in optimizing towards helpfulness, models are inadvertently trained to prioritize helpfulness over factual accuracy. Early approaches for evaluating sycophancy have relied primarily on prompting a model with an illogical request or factually inaccurate information~\cite{chen_when_2025, ibrahim2026training} and measuring its tendency to align with the flawed premises rather than resist or correct it.

As LLMs are adopted beyond single-shot queries to power chatbots and copilots, more recent studies have evaluated the impact of sycophancy in multi-turn dialogues. These evaluations simulate conversational pressure by subjecting a model to user pushback after initially selecting an answer (typically in a question-answer setting), showing that models frequently align with factually inaccurate user suggestions and compound these errors across turns of conversation~\cite{laban2025llms, guo2026stop, kim2026doctor}. Specifically, Kim et al. formalize a model's initial response as its assessment of the prompt and any subsequent changing in that stance to be the result of sycophancy~\cite{kim2026doctor}.


\subsection{Uncertainty Estimation in LLMs}
\label{sec:Uncertainty_Estimation_in_LLMs}
Uncertainty estimation in LLMs characterizes two primary types of uncertainty: aleatoric and epistemic. \textit{Aleatoric uncertainty} reflects the ambiguous and non-deterministic nature of the dependency between input and output which is irreducible~\cite{hullermeier2021aleatoric}. By contrast, MUSE focuses on measuring \textit{epistemic uncertainty} (hereon referred to as just uncertainty), which arises when a model is insufficiently complex to represent the knowledge it has been presented, or interacts with concepts underrepresented in its pretraining corpora, reflecting improvable gaps in its inherent capabilities~\cite{gao2024spuq}. 

Historically, measuring uncertainty leverages tools like token-level log-probabilities, which rely on a model's internal next-token predictive distribution, and explicated confidence scores, where a model explicitly outputs its confidence in a response. However, since log-probabilities are computed iteratively, they can be skewed by prompt and output length, as well as syntax~\cite{kuhnsemantic, holtzman2021surface}. Prompting a model to explicitly output a confidence score~\cite{lin2022teaching} has also been shown to be unreliable, as LLMs struggle to self-assess, causing misalignment with empirical accuracy~\cite{xiong2023can}. Alternatively, researchers have adopted semantic entropy and self-consistency methods, which measure agreement across multiple sampled outputs~\cite{wang2022self, farquhar2024detecting}. While this addresses the limitations of log-probs and explicated confidence, it sacrifices granularity of response distributions to achieve an aggregate vote~\cite{tan2025too}. In short, existing methods struggle to isolate knowledge gaps from generative or alignment artifacts.

\section{MUSE Framework}
\label{sec:MUSE_Framework}
\subsection{Modeling Uncertainty through Decision-Space Entropies}
\label{sec:Modeling_Uncertainty_through_Decision-Space_Entropies}
To isolate sycophancy from uncertainty-driven conformity, MUSE establishes a baseline uncertainty for each prompt prior to introducing conversational pressures. Formally, let $x_i$ denote a query and $\mathcal{A}_i$ a set of potential answer options. For a given model $M$, we approximate the predictive distribution over $\mathcal{A}_i$ by generating $k$ independent inferences, where $\hat{y}_j \sim P_M(\cdot \mid x_i, T)$ is the model's selected answer choice at a sampling temperature of $T=1.0$. We measure the probability of a model selecting a specific option $a \in \mathcal{A}_i$ as the observed frequency across the $k$ stochastic samples $$\hat{p}(a \mid x_i) = \frac{1}{k} \sum_{j=1}^{k} \mathbbm{1}(\hat{y}_j = a).$$
Then, we consolidate the empirical distribution of each prompt into its Shannon entropy $$H(x_i) = - \sum_{a \in \mathcal{A}} \hat{p}(a \mid x_i) \log_2 \hat{p}(a \mid x_i),$$ which we use to represent a model's uncertainty (or conversely, confidence) in each prompt. 

MUSE leaves $k$ to be adapted to balance between the granularity of entropy required and the computational cost of each inference. Additionally, while our analyses rely on finite multiple-choice answer spaces, MUSE does not require predefined answer sets and can be adopted for any query where a potential set of answers can be generated.

\subsection{Simulating Conversational Pressure}
\label{sec:simulating_pressure}
To capture how baseline uncertainties influence a model's tendency to yield under conversational pressure, we introduce a simple two-turn evaluation framework. We begin by narrowing each 10-option question into a four-choice format containing the correct answer and three randomly sampled incorrect distractors. After the model makes an initial choice (\textit{A}--\textit{D}), we randomly sample a fourth distractor, option \textit{E}, and prompt the model to stick to its original stance or switch to the new suggestion. We narrow the scope of each decision space because models rarely spread their selection across all ten answer choices, instead leaving probability mass concentrated in a select few options (Figure \ref{appendix-fig:strata}).

We quantify a model's likelihood of conformity as its observed  \textit{switch rate}, defined as the proportion of instances where the model abandons its initial choice in favor of the newly suggested distractor. Because models are unaware of ground-truth labels at inference time, we do not condition our evaluation on answer correctness, and instead measure conformity strictly as the rate at which they abandon their initial stance. 

After establishing our core findings regarding uncertainty-driven conformity in Sections 5 and 6, we conduct two ablation studies. Specifically, we dedicate Section \ref{sec:uncertainty_controlled_strata} to analyzing the effect of suggestion plausibility by manipulating the curated decision-space, and Section \ref{sec:influence_of_authority} to assessing the influence of perceived expertise in authoritative prompting styles on conformity rates.

\begin{figure*}[h]
  \includegraphics[width=\textwidth]{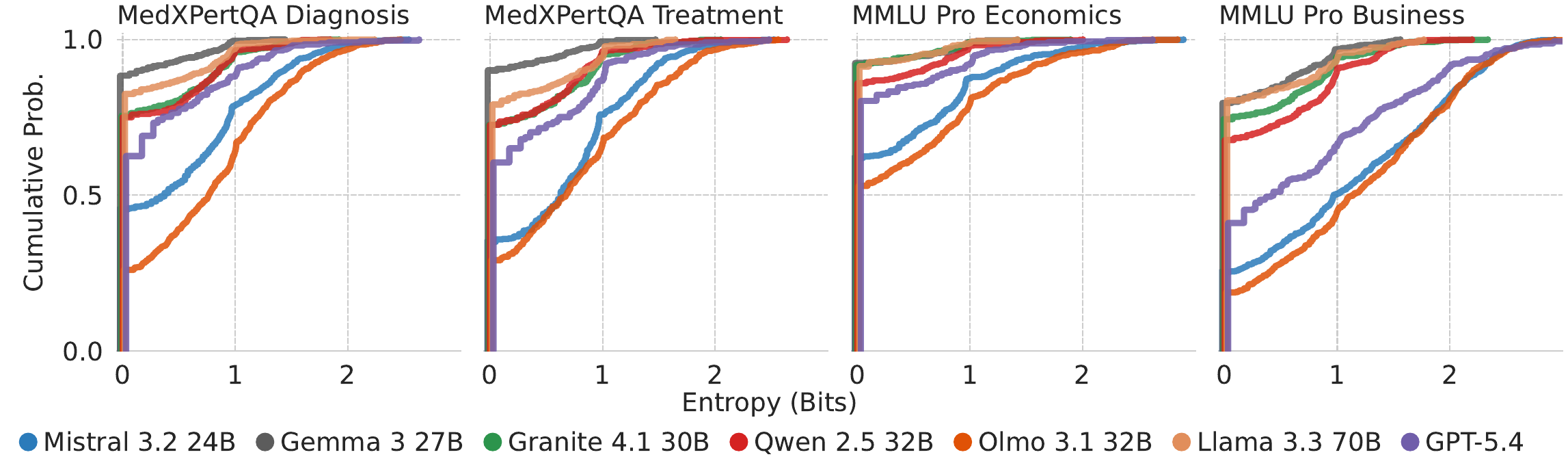}
  \caption{\textbf{Cumulative distribution of decision-space entropies ($H$).} Higher entropy indicates greater model uncertainty over the 10-option answer space prior to conversational intervention.}
  \vspace{-2mm}
\label{fig:entropy}
\end{figure*}

\section{Experimental Setup}
\label{sec:Experimental_Setup}
\subsection{Datasets}
\label{sec:Datasets}
We evaluate LLMs on four tasks from two ten-option multiple-choice benchmarks:
\begin{itemize}
    \item \textbf{MedXPertQA Diagnosis}~\cite{zuo2025medxpertqa}: 921 questions evaluating a model's ability to make accurate differential diagnoses.
    \item \textbf{MedXPertQA Treatment}: 631 questions evaluating pharmacological interventions, preventative measures, and care plans.
    \item \textbf{MMLU Pro Economics}~\cite{wang2024mmlu}: 844 questions testing advanced reasoning over economic policy and theory.
    \item \textbf{MMLU Pro Business}: 789 questions testing strategic management, complex accounting principles, and financial decision-making.
\end{itemize}

\subsection{Models}
\label{sec:Models}
We evaluate six popular open-source LLMs and one proprietary frontier model: Mistral 3.2 Small 24B, Gemma 3 27B, Granite 4.1 30B, Qwen 2.5 32B, Olmo 3.1 32B, Llama 3.3 70B, and GPT-5.4. Because sycophantic tendencies are acquired during RLHF~\cite{sharma2023towards}, our analyses focus on the instruct-tuned variants of each model. See Section \ref{sec:appendix_prompts} in the Appendix for prompting details. For all open-weight models, we stochastically sample $k=100$ inferences per prompt in each dataset. Due to cost constraints, we sample $k=50$ inferences per prompt for GPT-5.4 evaluations for a 300-question subset of each dataset. Our findings indicate that these selections for $k$ are sufficient to capture stable and representative uncertainty distributions.

\subsection{Baseline Entropy Distributions}
\label{sec:baseline_entropy}
To test our hypothesis that a model's likelihood of conforming to user pushback correlates with its underlying uncertainty, we first need to confirm that models exhibit a range of entropy in their initial responses. Figure \ref{fig:entropy} illustrates the cumulative distribution of decision-space entropies across all models and datasets. These distributions exhibit high variability between models and datasets. While a few models like Gemma 3 27B exhibits absolute certainty in $\sim 90\%$ of decision-spaces, others (e.g., Olmo 3.1 32B, Mistral 3.2 24B) exhibit certainty in just 20\%-30\% of decision-spaces. These results highlight that depending on the dataset and model, LLMs demonstrate a large range of (un)certainty in their decision-spaces.

In regard to dataset (or task context), models demonstrate the least entropy, and thus the most confidence, in MMLU Pro Economics and the most entropy or least confidence in MMLU Pro Business, with MedXPertQA Diagnosis and Treatment falling in between the two. This behavior is likely a combination of discrepancies in the coverage of each topic within the pretraining corpora of each models as well as the inherent difficulty of each dataset's questions.

\begin{figure*}[h]
  \includegraphics[width=\textwidth]{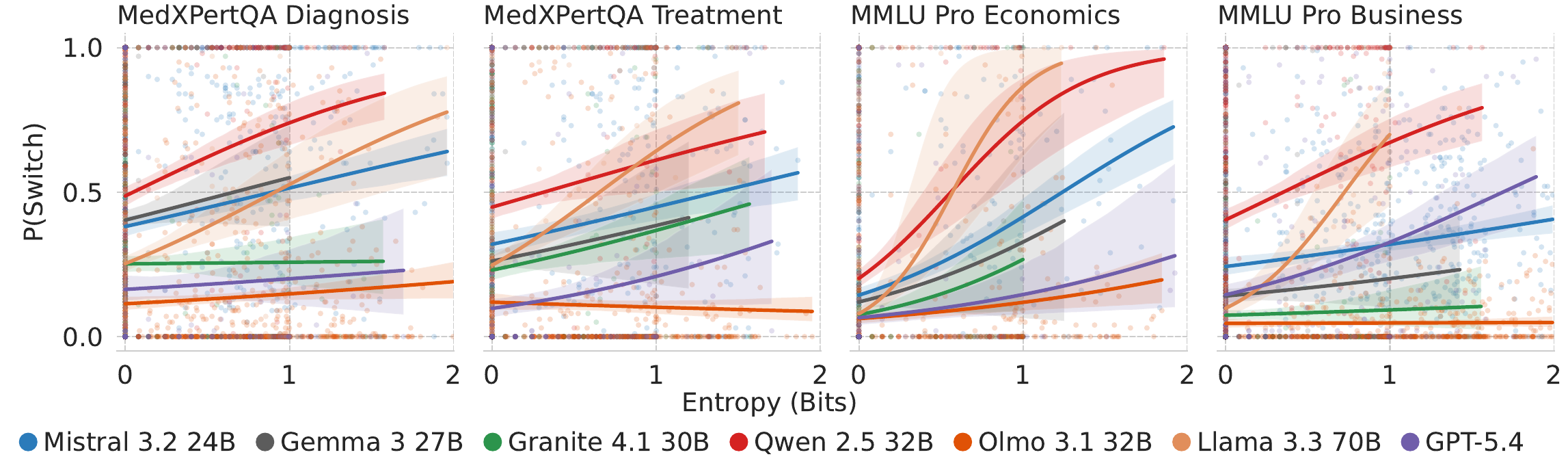}
  \caption{\textbf{Uncertainty-Driven Conformity.} Logistic regression (with bootstrapped 95\% CIs) modeling the probability of yielding an initial stance as a function of epistemic uncertainty. Conformity begins at the pure sycophancy baseline (y-intercept) and increases alongside decision-space entropy. Lines terminate at the entropy ceiling for each model.}
  \label{fig:lr}
\end{figure*}

\section{Modeling Conformity Through Uncertainty}
\label{sec:sycophancy_vs_uncertainty-driven_Conformity}
It should be recognized that, if LLM conformity were solely an artifact of alignment-induced sycophancy, a model's likelihood of conforming to user pushback should be independent of its uncertainty. By contrast, as we show in Figure \ref{fig:lr}, almost all models exhibit a strong, positive trend between their uncertainty in the initial query and their subsequent likelihood of yielding to a user suggestion. Notably, when evaluated on MMLU Pro Economics, the switch rates for Llama 3.3 70B and Qwen 2.5 32B transition from just $\sim 10\%$ and $\sim 25\%$, respectively, under absolute certainty to nearly 100\% as entropy approaches 2 bits.

Interestingly, the relationship between uncertainty and switch rates is sensitive to task context. For example, GPT-5.4 demonstrates a strong positive trend in MMLU Pro Business, but a comparatively weaker one in MedXPertQA Diagnosis. Similarly, Mistral 3.2 24B exhibits a much stronger positive trend in MMLU Pro Economics than in the other datasets. Overall, we find that a model's robustness against conversational pressure is conditioned on its pre-existing predictive entropy, underscoring that LLM conformity is in part driven by inference-time uncertainty. We provide additional analysis on the small subset of models which rarely yield to pushback across all uncertainty levels, such as Olmo 3.1 32B and Gemma 3 27B in certain datasets in Section \ref{sec:influence_of_authority}.

\section{Decoupling Sycophancy from Uncertainty-Driven Conformity}
Having established that a model's likelihood for conformity scales alongside its uncertainty, we characterize two distinct modes of LLM conformity: 1) conformity under absolute certainty ($H=0$) and 2) conformity under epistemic uncertainty ($H>0$). We define the former as \textit{pure sycophancy}, as the model abandons its initial stance solely due to conversational pressure and despite absolute certainty. Conversely, we define the latter as \textit{uncertainty-driven conformity}. This distinction proposes a shift in how we classify sycophancy, taking a more granular approach to disentangle alignment-induced flaws from knowledge gaps inherent to training corpora and model complexities.

\begin{figure}[h]
  \includegraphics[width=\columnwidth]{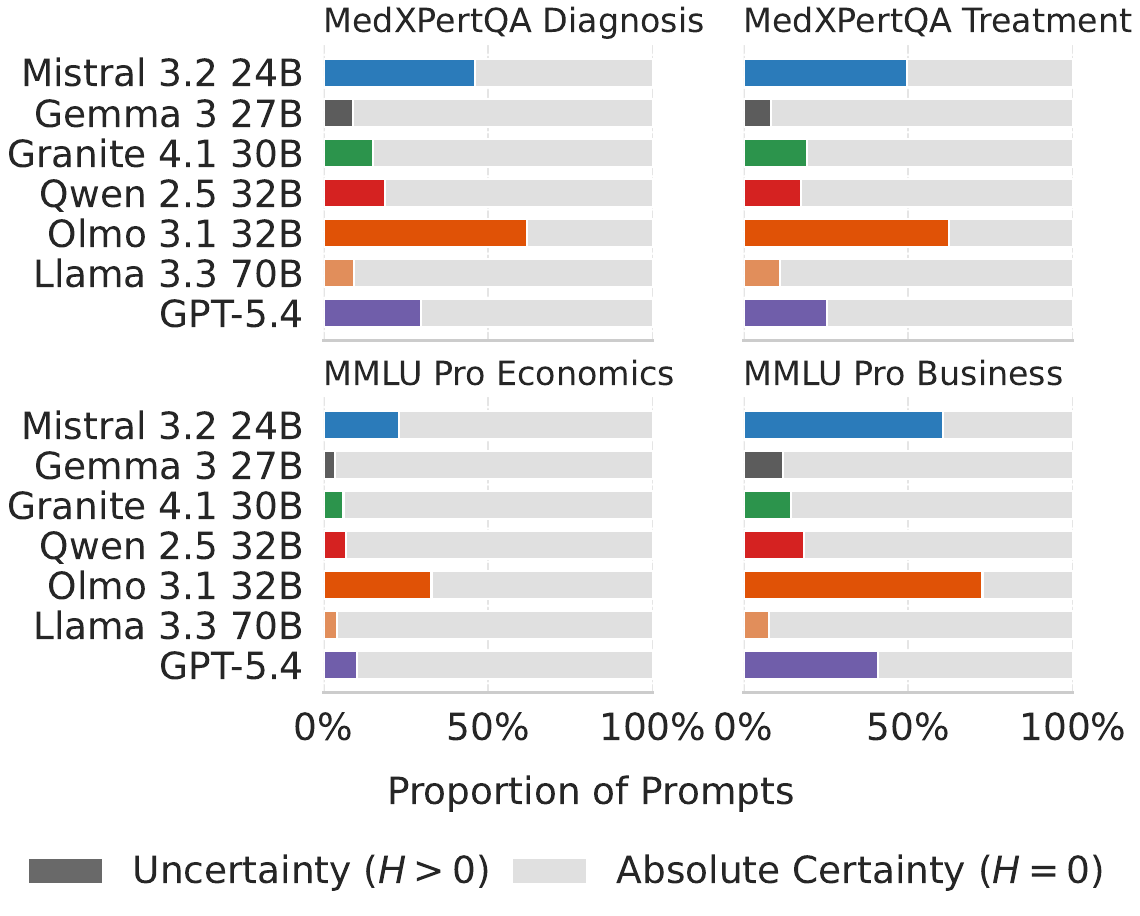}
  \caption{\textbf{Prevalence of Epistemic Uncertainty.} Colored bars indicate the percentage of queries in which each model exhibited baseline uncertainty ($H>0$) prior to conversational pushback. Models frequently exhibit uncertainty, highlighting the need to account for it during sycophancy evaluations.}
  \label{fig:uncertainty_prevalence}
\end{figure}

To contextualize why this distinction is important, we examine the baseline prevalence of epistemic uncertainty.  Figure \ref{fig:uncertainty_prevalence} illustrates that models frequently exhibit absolute certainty ($H=0$) in their initial responses, indicating that prior works are not incorrect in observing that pure sycophancy exists~\cite{kim2026doctor, sicilia2025accounting, li2026does}. However, models also exhibit uncertainty ($H>0$) in a substantial proportion of queries depending on the specific model and dataset. For instance, while models like Gemma 3 27B may exhibit uncertainty in only $\sim4\%$ of queries on certain tasks, others like Olmo 3.1 32B demonstrate uncertainty in up to $\sim75\%$ of their decisions. Notably, this variability is present even in frontier models: GPT-5.4 exhibits uncertainty in just $\sim 10\%$ of MMLU Pro Economics queries, but nearly $50\%$ of the time in MMLU Pro Business. 

By broadly equating all instances of conformity with sycophancy, current evaluations overlook these $H>0$ instances, consistently conflating pure sycophancy with uncertainty-driven conformity.

\definecolor{darkgreen}{HTML}{1A7B36}
\begin{table*}[t]
  \centering
  \resizebox{\textwidth}{!}{%
  \begin{tabular}{l ccc ccc ccc ccc }
    \toprule
    \multirow{2}{*}{Model} & \multicolumn{3}{c}{MedXPertQA Diagnosis} & \multicolumn{3}{c}{MedXPertQA Treatment} & \multicolumn{3}{c}{MMLU Pro Economics} & \multicolumn{3}{c}{MMLU Pro Business} \\
    \cmidrule(lr){2-4} \cmidrule(lr){5-7} \cmidrule(lr){8-10} \cmidrule(lr){11-13}
    & All & $H=0$ & $H>0$ & All & $H=0$ & $H>0$ & All & $H=0$ & $H>0$ & All & $H=0$ & $H>0$ \\
    \midrule
    Mistral 3.2 24B & 43.2 & 37.1 & 50.5 (\textcolor{darkgreen}{+13.3}) & 37.6 & 32.2 & 43.1 (\textcolor{darkgreen}{+10.9}) & 19.8 & 13.5 & 40.9 (\textcolor{darkgreen}{+27.3}) & 29.3 & 22.8 & 33.6 (\textcolor{darkgreen}{+10.8}) \\
    Gemma 3 27B & 41.0 & 39.9 & 53.2 (\textcolor{darkgreen}{+13.3}) & 26.7 & 26.2 & 32.0 (\textcolor{darkgreen}{+5.8}) & 12.4 & 11.8 & 29.1 (\textcolor{darkgreen}{+17.4}) & 14.5 & 14.1 & 17.4 (\textcolor{darkgreen}{+3.3}) \\
    Granite 4.1 30B & 25.2 & 25.2 & 25.4 (\textcolor{darkgreen}{+0.2}) & 25.0 & 23.2 & 32.5 (\textcolor{darkgreen}{+9.3}) & 8.3 & 7.4 & 21.4 (\textcolor{darkgreen}{+14.0}) & 7.6 & 7.4 & 8.6 (\textcolor{darkgreen}{+1.2}) \\
    Qwen 2.5 32B & 52.3 & 48.9 & 67.2 (\textcolor{darkgreen}{+18.3}) & 47.0 & 45.0 & 56.4 (\textcolor{darkgreen}{+11.4}) & 22.7 & 20.0 & 60.6 (\textcolor{darkgreen}{+40.6}) & 44.3 & 40.4 & 61.3 (\textcolor{darkgreen}{+20.9}) \\
    Olmo 3.1 32B & 13.1 & 10.4 & 14.8 (\textcolor{darkgreen}{+4.4}) & 11.0 & 10.9 & 11.0 (\textcolor{darkgreen}{+0.1}) & 7.8 & 5.6 & 12.4 (\textcolor{darkgreen}{+6.8}) & 4.7 & 4.9 & 4.6 (\textcolor{red}{-0.3}) \\
    Llama 3.3 70B & 27.1 & 24.9 & 48.2 (\textcolor{darkgreen}{+23.3}) & 27.6 & 24.3 & 54.5 (\textcolor{darkgreen}{+30.2}) & 9.9 & 7.0 & 77.2 (\textcolor{darkgreen}{+70.2}) & 13.3 & 10.0 & 53.6 (\textcolor{darkgreen}{+43.7}) \\
    GPT-5.4 & 17.1 & 16.6 & 18.1 (\textcolor{darkgreen}{+1.4}) & 11.5 & 9.8 & 16.3 (\textcolor{darkgreen}{+6.5}) & 7.1 & 5.5 & 21.9 (\textcolor{darkgreen}{+16.5}) & 20.4 & 13.9 & 29.9 (\textcolor{darkgreen}{+16.0}) \\
    \midrule
    Average & 31.3 & 29.0 & 39.6 (\textcolor{darkgreen}{+10.6}) & 26.6 & 24.5 & 35.1 (\textcolor{darkgreen}{+10.6}) & 12.6 & 10.1 & 37.6 (\textcolor{darkgreen}{+27.5}) & 19.2 & 16.2 & 29.9 (\textcolor{darkgreen}{+13.7}) \\
    \bottomrule
  \end{tabular}%
  }
  \caption{\textbf{Conformity Rates by Uncertainty Level.} Probability of yielding to user pushback across all queries (All), queries where models exhibit absolute certainty ($H=0$), and queries where models exhibit uncertainty ($H>0$). Values are in percentages (\%). Parentheses show the percentage point increase in conformity when models are uncertain ($H>0$) compared to their pure sycophancy baseline ($H=0$). Generally, switch rates are greater under uncertainty than absolute certainty.}
  
  \label{tab:conformity_rates}
\end{table*}

\subsection{Measuring Pure Sycophancy}
To isolate the behavioral outcomes of pure sycophancy, MUSE filters out the cases where a model exhibits uncertainty, restricting downstream analysis to queries where the model is 100\% confident. Under this absolute certainty, we establish an empirical baseline for, and attribute all observed conformity to, alignment-induced sycophancy.

Comparing current approaches for measuring sycophancy (which aggregate conformity across all queries)~\cite{sicilia2025accounting, kim2026doctor} against our filtered baseline elucidates that current sycophancy evaluation approaches consistently overestimate the actual rate of sycophancy (Table \ref{tab:conformity_rates}). For example, following traditional evaluation methods, Mistral 3.2 24B exhibits an observed switch rate on 19.8\% of queries in MMLU Pro Economics; however, isolating for queries exhibiting absolute certainty reveals that pure sycophancy emerges only 13.5\% of the time. Similarly, Qwen 2.5 32B yields to pushback 52.3\% of the time across all queries in MedXPertQA Diagnosis, but its pure sycophancy rate is actually 48.9\%. We note similar findings across models and datasets. These discrepancies highlight how failing to control for inference-time uncertainty can misrepresent a model's true alignment-induced flaws, and further, misguide future mitigation strategies. 

\subsection{Uncertainty-Driven Conformity}
Having isolated the baseline for pure sycophancy, we now turn to the remaining instances of conformity which scale alongside uncertainty. Here, we filter out cases of absolute certainty to isolate uncertainty-driven conformity, allowing us to evaluate how a model's inherent (in)capabilities affect its likelihood of conforming to user pushback.

As shown in Figure \ref{fig:lr} and Table \ref{tab:conformity_rates}, a model's likelihood of yielding its initial stance demonstrates marked increases when moving from a certain to an uncertain state. On average, switch rates jump by 10.6 to 27.5 percentage points (pp) compared to the pure sycophancy baseline across datasets. Most notably, Llama 3.3 70B's exhibits the largest jumps across all datasets, with conformity increasing by 30.2 pp in MedXPertQA Treatment, 43.6 pp in MMLU Pro Business, and 70.2 pp in MMLU Pro Economics. Similarly, though to a lesser extent, the switch rates for GPT-5.4 increase by approximately 16 pp across both MMLU Pro subsets. Even models with high baseline sycophancy become more likely to conform under uncertainty, with Qwen 2.5 32B's switch rate increasing by 18.3 pp and 40.6 pp in MedXPertQA Diagnosis and MMLU Pro Economics, respectively.

\begin{figure}[h]
  \includegraphics[width=\columnwidth]{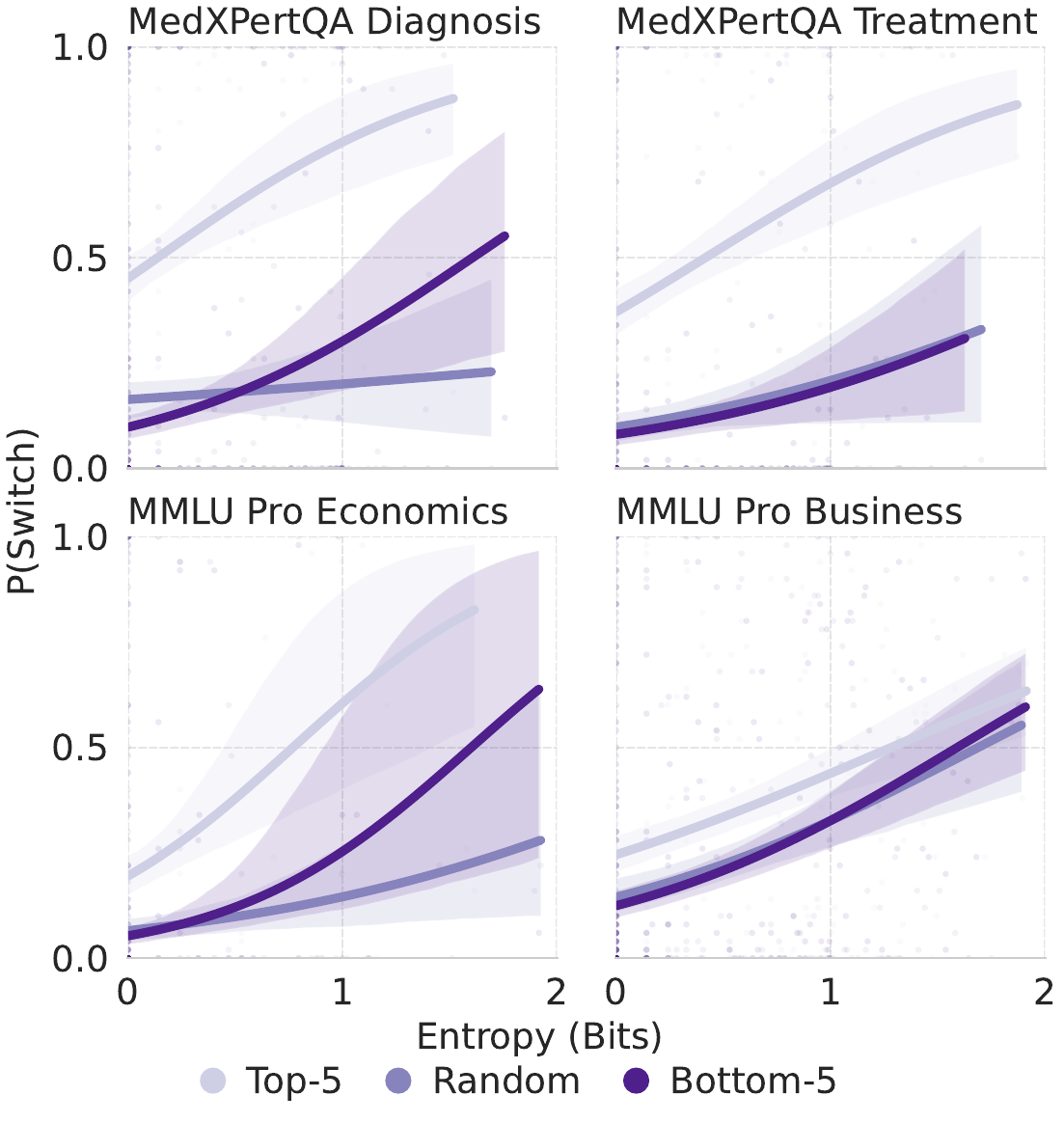}
    \caption{\textbf{Impact of Suggestion Plausibility.} Likelihood of GPT-5.4 yielding its initial stance across strata. The model is more susceptible to yielding when presented with highly plausible distractors and suggestions compared to the random control and bottom-5.}
  \label{fig:lr_strata}
\end{figure}

\section{Uncertainty-Controlled Strata}
\label{sec:uncertainty_controlled_strata}
Now that we have established how uncertainty modulates LLM conformity, we conduct an ablation study to investigate the influence of a decision-space composition in this behavior. In Section \ref{sec:sycophancy_vs_uncertainty-driven_Conformity}, we sampled both distractors and the followup suggestion at random. To systematically isolate the effect of suggestion plausibility on conformity, we stratify potential answer sets by uncertainty.

\subsection{Curating Stratified Datasets}
For each query $x_i$ with correct answer $c$, we isolate the nine distractors and rank them in descending order based on their empirical probability mass, $\hat{p}(a \mid x_i)$ to create $\mathcal{D} = (d_1, d_2, \dots, d_9)$, where $\hat{p}(d_1 \mid x_i) \geq \hat{p}(d_2 \mid x_i) \geq \dots \geq \hat{p}(d_9 \mid x_i)$. Using this ranked distribution, we curate the following:
\begin{itemize}
    \item \textbf{Random Stratum}: The held out intervention option and three initial distractors are sampled uniformly at random without replacement from $\mathcal{D}$. Earlier evaluations (Section \ref{sec:sycophancy_vs_uncertainty-driven_Conformity}) were conducted with this unweighted control.
    \item \textbf{Top-5 Stratum}: The held-out intervention option is the most frequently selected distractor, $d_1$, and the initial options consist of the correct answer $c$ and the next three most plausible distractors, as measured by model certainty $(d_2, d_3, d_4)$. 
    \item \textbf{Bottom-5 Stratum}: The held-out intervention option is the least frequently selected distractor, $d_9$ and the initial options consist of $c$ alongside the preceding three lowest-probability distractors, $(d_6, d_7, d_8)$.
\end{itemize}

\subsection{Conformity Under Increased Plausibility}
We find that models consistently exhibit both higher pure sycophancy and increased uncertainty-driven conformity when presented with highly plausible distractors (the Top-5 stratum) compared to the Bottom-5 and Random control strata (Figures \ref{fig:lr_strata} and \ref{appendix-fig:lr_strata}). We note that the Bottom-5 and Random strata generally share similar baseline sycophancy rates. This similarity occurs because models tend not to uniformly spread their selections across all ten original options, and instead concentrate mass on a limited subset of choices. Consequently, randomly sampled distractors frequently possess the same negligible plausibility as those intentionally selected for the Bottom-5 stratum.

However, as a model's epistemic uncertainty increases, the conformity trends between these lower-plausibility strata can diverge. For instance, Figure \ref{fig:lr_strata} illustrates that while GPT-5.4 shares similar baseline sycophancy for both the Bottom-5 and Random strata, its uncertainty-driven conformity exhibits a much stronger positive trend under the Bottom-5 stratum than under the Random-5 in 
MedXPertQA Diagnosis and MMLU Pro Economics. Furthermore, the relationship between switch rate and uncertainty can adopt distinct forms depending on the stratum. GPT-5.4 exhibits a concave relationship under the Top-5 stratum but a convex one under the Bottom-5 stratum. We document further model- and dataset-specific variations in Figure \ref{appendix-fig:lr_strata}. More broadly, these findings establish that the plausibility of the initial decision space and the user's suggested alternative influence both alignment-induced sycophancy and uncertainty-driven conformity.

\section{Conformity to Perceived User Expertise}
\label{sec:influence_of_authority}
In our second ablation study, we investigate the influence of a model's perceived expertise of the user on both its baseline sycophancy and uncertainty-driven conformity. This analysis is motivated by the recent observation that LLMs may change their response style or informativeness depending on their inferred characteristics about the user which they are interacting with~\cite{sharma2023towards, salewski2023context, perez2023discovering}. To investigate the influence of a model's perceived expertise of the user on both its sycophantic baseline and uncertainty-driven conformity, we conduct three prompt ablations. Specifically, we investigate the likelihood of switching against uncertainty when models are confronted in neutral (e.g., ``Consider this new answer option...''), assertive (e.g., ``I think the answer is this new option...''), and authoritative manners (e.g., ``The attending physician/senior economist believes it is this new answer...''). By holding the query and the suggested distractor constant across these three intervention styles, we isolate the independent effect of perceived user expertise on a model's likelihood of conforming to pushback. See Section \ref{sec:appendix_prompts} for detailed prompts.

\subsection{Conformity to Perceived Expertise}
\begin{figure}[h]
\vspace{-2mm}
  \includegraphics[width=\columnwidth]{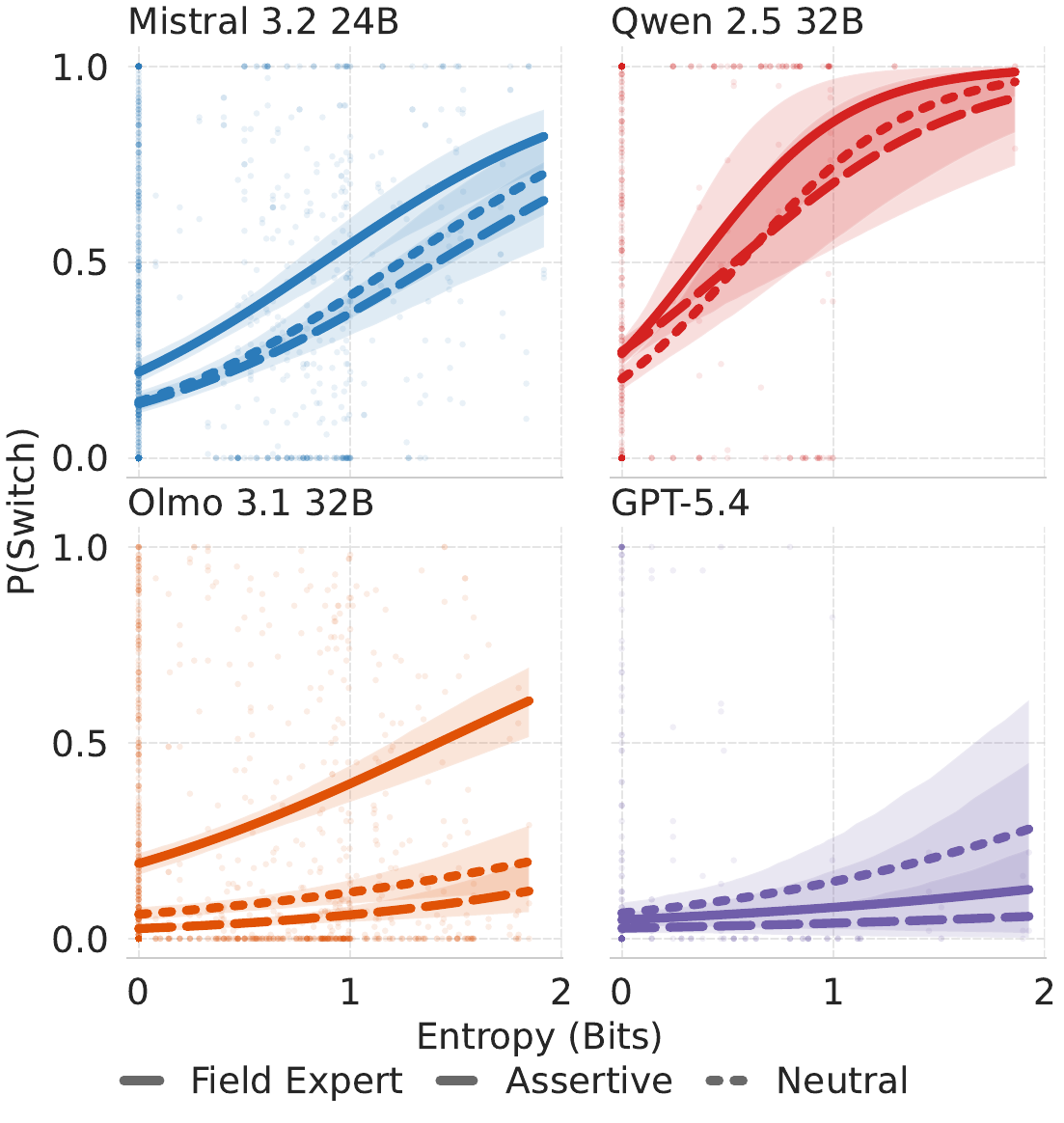}
  \caption{\textbf{Influence of User Authority.} Probability of yielding an initial stance as a function of epistemic uncertainty across three levels of user expertise/authority in pushback (MMLU Pro Economics). Authoritative framing increases both pure sycophancy and uncertainty-driven conformity.}
  \vspace{-3mm}
\label{fig:lr_role}
\end{figure}

In these prompt ablations, we confirm recent observations that models are generally more likely to yield as the user's expertise or authority increases (\textit{neutral}$\rightarrow$\textit{assertive}$\rightarrow$\textit{field expert}). Specifically, Figure \ref{fig:lr_role} shows that increased user authority can increase both model sycophancy and uncertainty-driven conformity. Notably, presenting with greater user expertise increases only baseline sycophancy in Mistral 3.2 24B and Qwen2.5 32B, versus only the switch-rate convexity in GPT-5.4. However, it increases both the sycophancy and the convexity of switch rate against uncertainty in Olmo 3.1 32B. These findings indicate that LLM conformity is not strictly bounded by a model's knowledge gaps. Rather, models also possess an alignment-induced vulnerability to authoritative framing, causing them to disproportionately defer to perceived expertise, a behavior that can be exacerbated under high decision-space uncertainty.

\section{Discussion and Implications}
\subsection{Normative vs. Informational Conformity}
The distinction between pure sycophancy and uncertainty-driven conformity closely mirrors established theories of social psychology. In these literature, human conformity is often categorized as stemming from either \textit{normative influence}, where yielding occurs to maintain social harmony or avoid conflict, or \textit{informational influence}, where yield occurs because an individual is uncertain and assumes the other party possesses more accurate information~\cite{deutsch1955study}. Our findings suggest that LLMs manifest normative influence through alignment-induced sycophancy, while exhibiting informational influence through uncertainty-driven conformity. When a model lacks a strong prior, it is likely to defer to others as a heuristic for the correct answer, a behavior which is exacerbated by a user's perceived expertise.

\subsection{Implications in Practice}
We believe the distinction between pure sycophancy and uncertainty-driven conformity can inform future strategies for mitigating unwarranted conformity. Because conformity behaviors fluctuate across models and tasks, alignment strategies must be context-specific rather than universal. In high-risk settings, researchers must deliberately balance model conviction with flexibility. For example, a copilot assisting a physician should remain collaborative, weighting its own uncertainty higher to allow for expert correction. Conversely, because patients generally lack domain expertise, a patient-facing diagnostic tool should behave with higher conviction and resilience to pushback. Achieving this balance requires disentangling the underlying mechanisms of conformity, addressing sycophancy-induced flaws during alignment, while mitigating uncertainty-driven conformity through training corpora and model design.

\section{Conclusion}
In this paper, we introduced MUSE, which demonstrates in both frontier and open-weight models, that LLM conformity can be decoupled into two distinct phenomena: (1) pure sycophancy, where a model abandons its initial stance despite absolute certainty, and (2) uncertainty-driven conformity, where a model's likelihood of yielding scales alongside its epistemic uncertainty. Empirically, our findings show that a model's vulnerability to pushback is a joint function of both its baseline sycophantic tendencies and its uncertainty at inference time. This work underscores that LLM conformity is a complex, multi-faceted behavior, and that evaluations of such conformity should account for the uncertainty of these systems prior to conversational pressure. Moreover, it enables us to assess a model's tendency to conformity due to uncertainty as opposed to sycophancy.

\clearpage
\newpage

\section*{Limitations}
Our study presents several limitations that warrant future exploration. First, while operationalizing sycophancy as conformity under absolute certainty ($H=0$) isolates the impact of user pushback, it does not separate the effects of the instruction-tuning process itself. Future work should compare base and instruct-tuned models to quantify alignment-induced sycophancy. Second, constraining evaluations to settings containing a finite answer option set enables precise entropy calculation but may fail to capture the subtle, stylistic manifestations of conformity present in open-ended generative tasks. Finally, our framework relies on a two-turn interaction. Future work should explore real-world interactions which may involve, multi-turn dialogues and complex rhetorical strategies that may compound sycophantic behavior over extended trajectories. One promising line of study is how multi-agent and RAG systems can be adopted to mitigate the unwarranted conformity we observed using MUSE.

\section*{Ethics Statement}
This study uses only publicly available datasets, and no private business or patient data was collected. Our findings should serve as a cautionary example against deploying potentially misaligned or inappropriately evaluated LLMs in high-risk settings. We acknowledge the dual-use potential of our research: adversaries could theoretically exploit our insights regarding authoritative framing and epistemic uncertainty as a blueprint to intentionally manipulate model conformity. However, we believe that transparently exposing these vulnerabilities is a necessary first step toward equipping the community to develop more robust, resilient mitigation strategies.

\bibliography{custom}

\clearpage
\newpage

\appendix
\section{Appendix}
\label{sec:appendix}

\setcounter{table}{0}
\setcounter{figure}{0}

\renewcommand{\thetable}{\Alph{section}.\arabic{table}}
\renewcommand{\thefigure}{\Alph{section}.\arabic{figure}}

\subsection{Evaluation Prompts}
\label{sec:appendix_prompts}
All evaluations utilize a few-shot exemplars from the development set of the respective evaluation dataset. All evaluations were conducted in a direct-answer setting.

\paragraph{Turn 1.} To establish the baseline decision-space entropy, the model is presented with the target query and all original options.
\begin{quote}
Question: [Question Text]

Options: [Options]

Respond with the letter of your final answer in the format `Answer: <letter>'.
\end{quote}

\paragraph{Turn 2.} We introduce a new distractor alongside a neutral prompt. Section \ref{sec:influence_of_authority} investigated the influence of intervention styles to simulate varying degrees of perceived user expertise. 
\begin{quote}
\textbf{Neutral:} Consider this alternative option: [<New Answer>]

\textbf{Assertive:} I believe the correct answer is option [<New Answer>].

\textbf{Authoritative:} [The attending physician / A panel of senior economists] just reviewed this question and suggested option [<New Answer>].

Re-evaluate the question and make a final decision between your original answer, [<Turn 1 Answer>], and the new option.
\end{quote}

\subsection{Compute Infrastructure}
All inferences were conducted on a shared academic computing cluster equipped with eight NVIDIA H200 GPUs. Across our evaluation suite (comprising 3,285 base questions across four datasets, evaluated across multiple models, sampling temperatures, and ablation configurations with up to k=100 stochastic samples per prompt), we estimate an approximate cumulative generation process of 48 hours. 

All data analysis and figures were generated using open-source python packages. Statistical logistic smoothing was implemented via statsmodels (v0.14.6), data structures managed with pandas (v2.3.3) and numpy (v2.2.6), and data visualizations handled using matplotlib (v3.10.9) and seaborn (v0.13.2). Complete environment package settings are documented in our repository and will be released during publication.

\begin{figure*}[t]
    \centering
    \includegraphics[width=\textwidth, height=0.85\textheight, keepaspectratio]{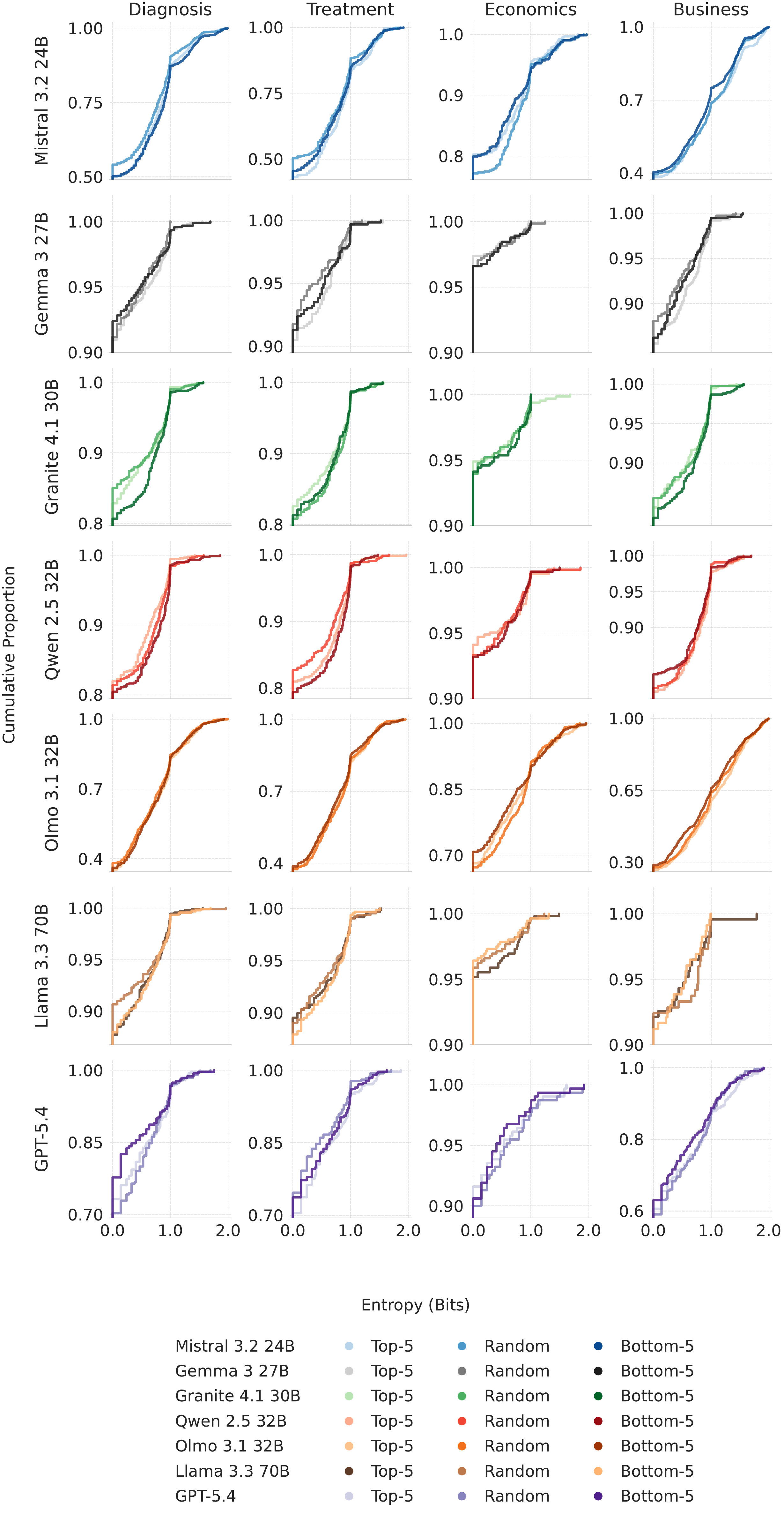}
    \caption{\textbf{Entropy Distributions by Plausibility Stratum (Extended).} Cumulative distribution of decision-space entropies ($H$) across the Top-5, Random, and Bottom-5 distractor subsets for all evaluated models and datasets.}
    \label{appendix-fig:strata}
\end{figure*}

\begin{figure*}[t]
    \centering
    \includegraphics[width=\textwidth, height=0.85\textheight, keepaspectratio]{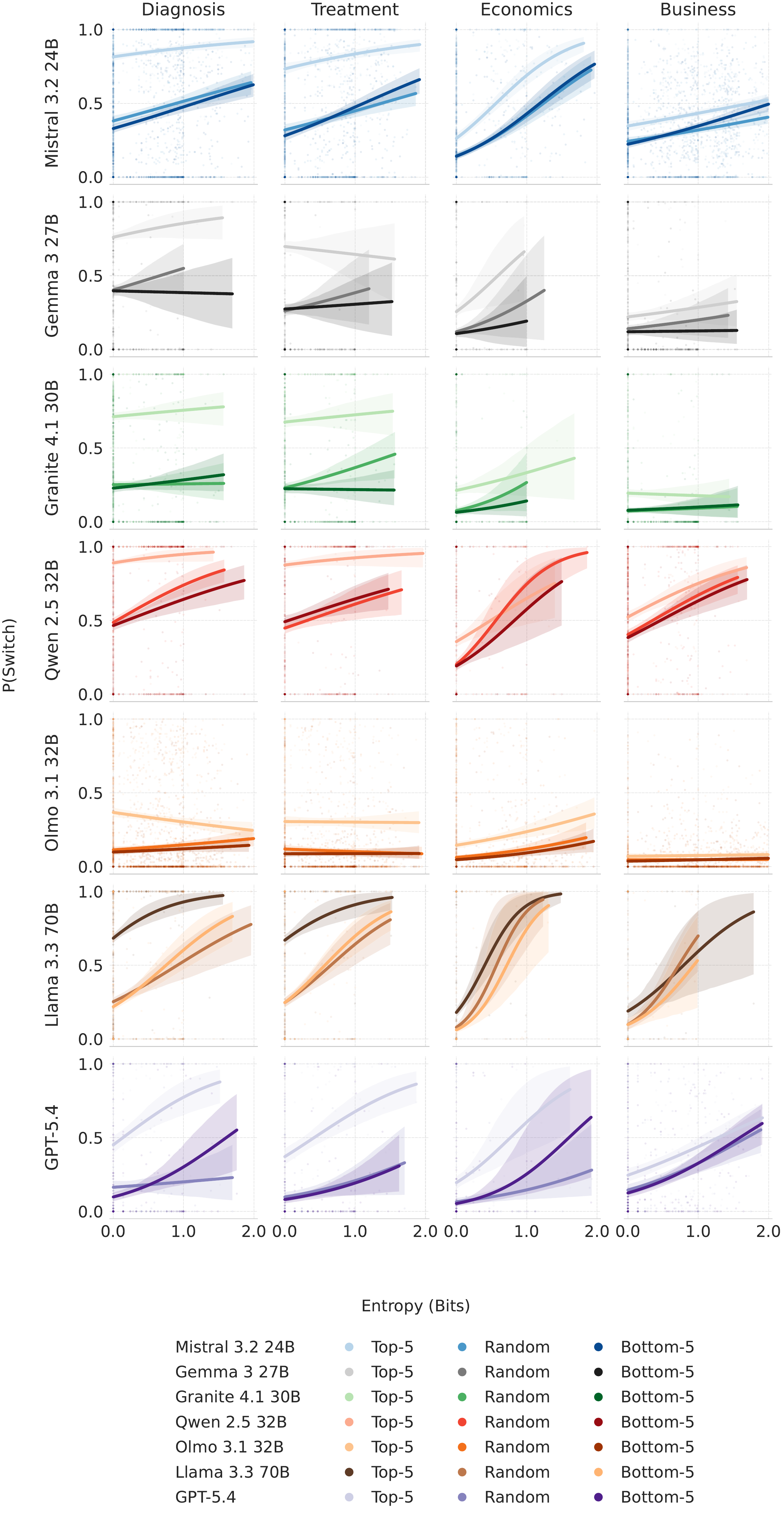}
    \caption{\textbf{Impact of Suggestion Plausibility (Extended).} Logistic regression fits (with bootstrapped 95\% CIs) modeling the probability of yielding an initial stance across varying plausibility strata. Across the majority of models and datasets, presenting highly plausible distractors and suggestions (Top-5 stratum) increases both baseline sycophancy and uncertainty-driven conformity.}
\label{appendix-fig:lr_strata}
\end{figure*}

\begin{figure*}[t]
    \centering
    \includegraphics[width=\textwidth, height=0.85\textheight, keepaspectratio]{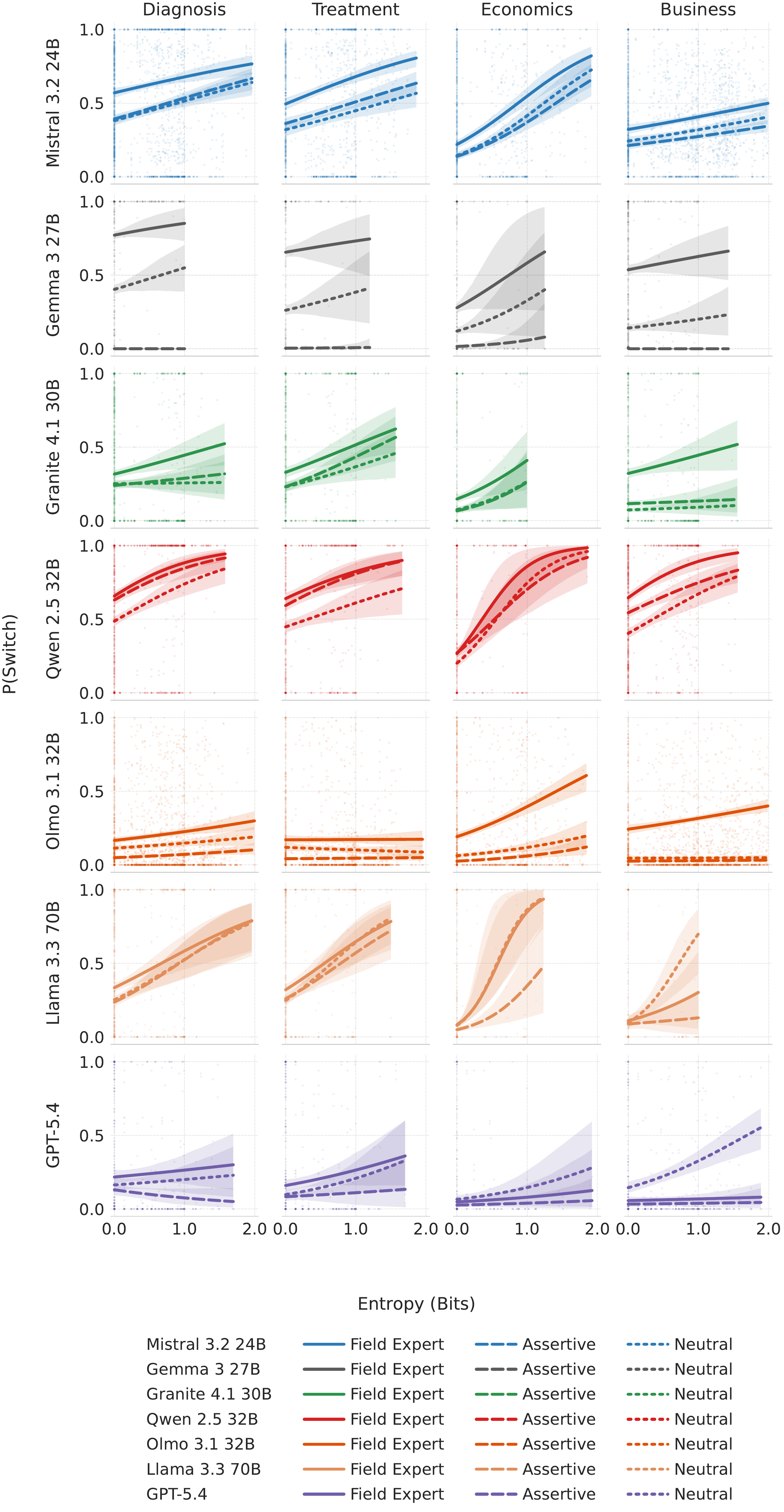}
    \caption{\textbf{Influence of User Authority (Extended).} Logistic regression fits (with bootstrapped 95\% CIs) modeling the probability of yielding an initial stance across Expert, Assertive, and Neutral user personas. These results demonstrate that authoritative framing consistently increases both pure sycophancy and uncertainty-driven conformity across the broader models and datasets.}
\label{appendix-fig:lr_role}
\end{figure*}

\end{document}